\documentclass[a4paper,12pt]{spieman}  
\usepackage{amsmath,amsfonts,amssymb}
\usepackage{microtype}
\usepackage{graphicx,epsfig}
\usepackage[caption=false,font=footnotesize]{subfig}
\usepackage{tabularx}
\usepackage[linesnumbered,boxed]{algorithm2e}
\usepackage{setspace}
\def\F{\mathbf{F}}

\usepackage{color}

\title{Artifact reduction for separable non-local means}
\author{Sanjay~Ghosh and Kunal~N.~Chaudhury\footnote{Email: sanjayg@iisc.ac.in, kunal@iisc.ac.in}}
\affil{Department of Electrical Engineering, Indian Institute of Science, Bangalore.}

\begin{document}

\maketitle
\begin{abstract}
It was recently demonstrated [J. Electron. Imaging, \textbf{25}(2), 2016] that one can perform fast non-local means
(NLM) denoising of one-dimensional signals using a method called lifting. The cost of lifting is independent of the patch length, which dramatically
reduces the run-time for large patches. Unfortunately, it is difficult to directly extend lifting for non-local means denoising of images. To bypass this, the authors
proposed a separable approximation in which the image rows and columns are filtered using lifting. The overall algorithm is significantly faster than
NLM, and the results are comparable in terms of PSNR. However, the separable processing often produces vertical and horizontal stripes in the
image. This problem was previously addressed by using a bilateral filter-based post-smoothing, which was effective in removing some of the
stripes. In this letter, we demonstrate that stripes can be mitigated in the first place simply by involving the neighboring rows (or columns) in
the filtering. In other words, we use a two-dimensional search (similar to NLM), while still using one-dimensional patches (as in the previous
proposal). The novelty is in the observation that one can use lifting for performing two-dimensional searches. The proposed approach produces
artifact-free images, whose quality and PSNR are comparable to NLM, while being significantly faster.
\end{abstract}

\keywords{Denoising, non-local means, fast algorithm, lifting, artifact}

\pagestyle{myheadings}
\thispagestyle{plain}
\markboth{}{}

\section{Introduction}
\label{introduction}

We consider the problem of denoising grayscale images corrupted with additive white Gaussian noise. A popular denoising method is the non-local means (NLM) algorithm \cite{Buades2005}, where image patches are used to perform pixel aggregation.
While NLM is no longer the state-of-the-art, it is still used in the image processing community due to its simplicity, decent denoising performance, and the availability of fast implementations.
The NLM of an image $f = \{f(i) : i \in \Omega\}$, where $\Omega  = \big\{ i=(i_1, i_2):  1\leq i_1,i_2 \leq N\}$, is given by \cite{Buades2005}
\begin{equation}
\label{NLM}
\text{NLM}[f](i)  = \frac{\sum_{j \in S(i)} w_{ij} f(j)}{\sum_{j \in S(i)} w_{ij} } \qquad (i \in \Omega).
\end{equation}
where $S(i)=i+[-S,S]^2$ is a search window around the pixel of interest. The weights $w_{ij}$ are set to be
\begin{equation}
\label{wtsNLM}
 w_{ij} = \exp \Big(\! - \frac{1}{\alpha^2} \sum_{k \in P} \big( f(i+k) - f(j+k) \big)^2 \Big),
\end{equation}
where $\alpha$ is a smoothing parameter and $P= [-K,K]^2$ is a two-dimensional patch.

A direct implementation of \eqref{NLM} has the per-pixel complexity of $O(S^2K^2)$, where $S$ and $K$ are typically in the range $[7,20]$ and $[1,3]$.\cite{Buades2005}
Several computational tricks and approximations have been proposed to speedup the direct implementation. \cite{Sapiro2005, Wang2006,Darbon2008,Dauwe2008,Orchard2008,KUD2009, Condat2010}. A particular means to speed up NLM is using a separable approximation, which in fact is a standard trick in the image processing literature \cite{N1981,Pham2005,KLCLKK2011,Fukushima2015}. In separable filtering, the rows are processed first followed by the columns (or in the reverse order). Of course, if the original filter is non-separable, then the output of separable filtering  is generally different from that of the original filter, since a natural image typically contains diagonal details \cite{KLCLKK2011}. This is the case with NLM since expression \eqref{wtsNLM} is not separable.
The present focus is on a recent separable approximation of NLM.\cite{Ghosh2016} At the core of this proposal is a method called lifting, which computes the NLM of a one-dimensional signal using $O(S)$ operations per sample. In other words, the complexity of lifting is independent of the patch length $K$. Extending lifting for NLM denoising of images, however, turns out to be a difficult task. Therefore, we proposed a separable approximation, called separable NLM (SNLM)\cite{Ghosh2016}, in which the rows and columns of the image are independently filtered using lifting. In particular, we separately computed the ``rows-then-columns'' and ``columns-then-rows'' filtering, which were then optimally combined. The per-pixel complexity of SNLM is $O(S)$, which is a dramatic reduction compared to the $O(S^2K^2)$ complexity of NLM.
 
 A flip side of SNLM (as is the case with other separable formulations\cite{Gastal2011}) is that often vertical and horizontal stripes are induced in the processed image.  
The stripes are more prominent along the last filtered dimension.\cite{Gastal2011} In SNLM, this problem was alleviated using the optimal recombination mentioned above followed by a bilateral filter-based post-smoothing. In this work, we demonstrate that the stripes can be mitigated in the first place simply by involving the neighboring rows (or columns) in the filtering. 
In other words, we use a two-dimensional search (similar to classical NLM\cite{Buades2005}), while still using one-dimensional patches (as done previously \cite{Ghosh2016}).
The present novelty is in the observation that one can use lifting for performing a  two-dimensional search. In particular, the per-pixel complexity of the proposed approach is  $O(S^2)$, which is higher than our previous proposal, but still substantially lower than that of classical NLM. Importantly, the proposed approach no longer exhibits the visible artifacts that are otherwise obtained using SNLM.

The rest of the paper is organized as follows. We recall the SNLM algorithm in Section \ref{sec:prior works} and its fast implementation using lifting. We also illustrate the artifact problem with an example. The proposed solution is presented in Section \ref{sec:proposed}, along with some algorithmic details. In Section \ref{sec:results}, we report the denoising performance of our approach and compare it with classical NLM and SNLM. We end the paper with some concluding remarks in Section \ref{sec:conclusion}.

\begin{figure}[!h]
\centering
 \includegraphics[width=0.97\linewidth]{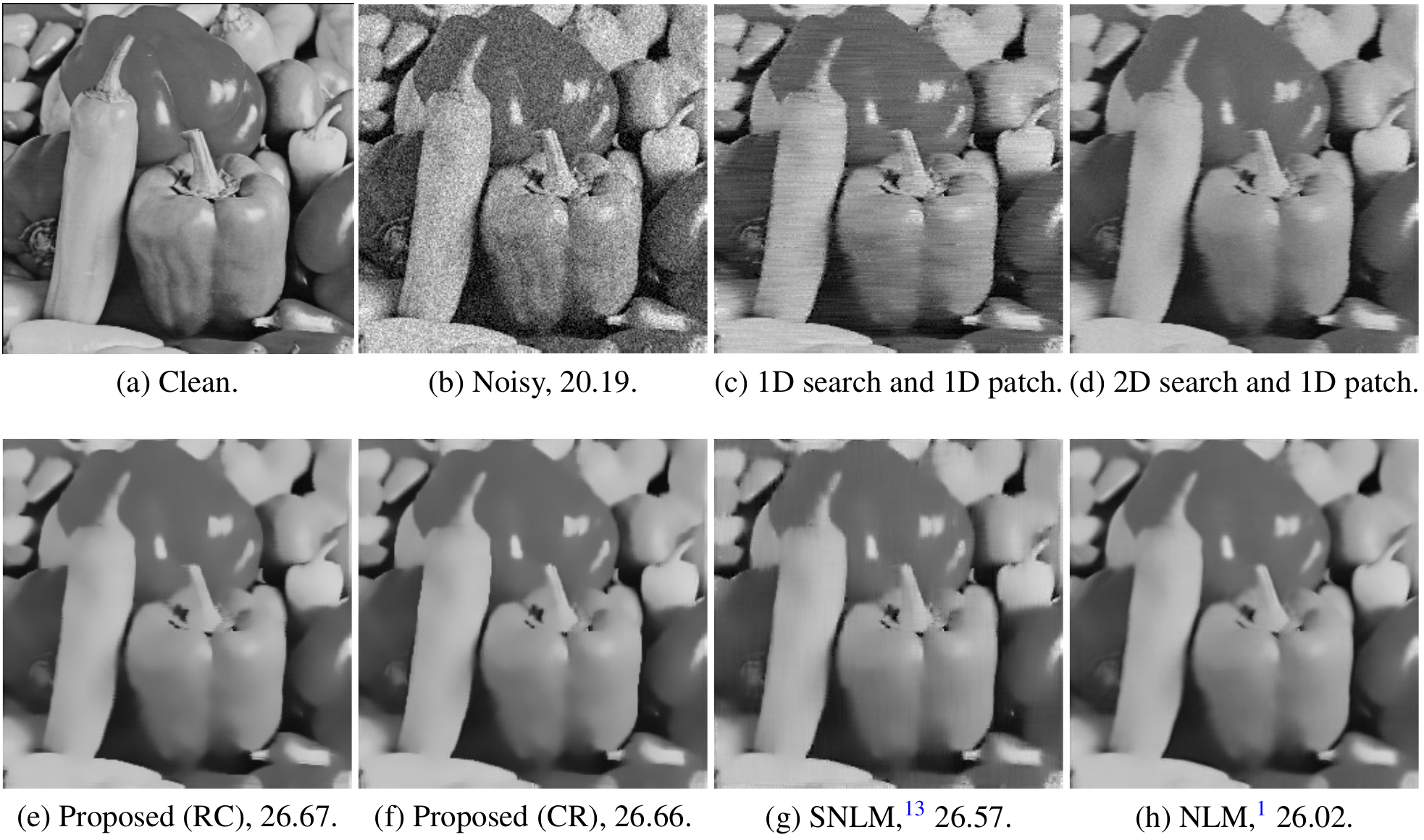}
\caption{Denoising of \emph{Peppers} at noise standard deviation $\sigma=25$.
We see stripes in (c) in which both the patch and search window are one-dimensional (both are along rows).
As seen in (d), the stripes can however be reduced using a two-dimensional search in place of the one-dimensional counterpart (though we still see some noise).
The image obtained by further processing (d) using a two-dimensional search and one-dimensional patches (along columns) is shown in (e).
The visual quality and PSNR (mentioned below each image) of (e) is comparable to NLM.
In (f), we have reversed the order of processing: we first use one-dimensional patches along columns and then along rows (the search is two-dimensional). Notice that the order (RC/CR) has no visible impact on the final output. 
Also notice that residual stripes can be seen in SNLM.}
\label{example}
\end{figure}

\section{Separable Non-Local Means}
\label{sec:prior works}

To set up the context, we briefly recall the SNLM algorithm\cite{Ghosh2016}. Suppose we have a one-dimensional signal $g=\{g(i) : 1 \leq i \leq N\}$, corresponding to a row or column. 
The one-dimensional analogue of \eqref{NLM} is given by
\begin{equation}
\label{NLM1D}
\text{NLM$1$D}[g](i)  = \frac{\sum_{j \in S(i)} w_{ij} g(j)}{\sum_{j \in S(i)} w_{ij} } \qquad (1 \leq i \leq N),
\end{equation}
and
\begin{equation}
\label{weights}
 w_{ij} = \exp \Big(\! - \frac{1}{\beta^2} \sum_{k=-K}^K \big( g(i+k) - g(j+k) \big)^2 \Big),
\end{equation}
where $S(i)=i+[-S,S]$ and $\beta$ is a smoothing parameter. In other words, both the search window and patch are one-dimensional in this case. It was observed in our previous work that the weights $\{w_{ij}: 1 \leq i \leq N, i-S \leq j \leq i+S\}$ can be computed using $O(1)$ operations with respect to $K$. In particular, consider the $N \times N$ matrices:
\begin{equation}
\label{tensor}
\F(i,j) = g(i) g(j) \quad \quad  (1 \leq i,j \leq N),
\end{equation}
and
\begin{equation}
\label{convn}
\overline{\F}(i,j) =  \sum_{k=-K}^K  \F(i + k,j + k).
\end{equation}
\\
We see that $\overline{\F}$ is the smoothed version of $\F$, obtained by box filtering $\F$ along its sub-diagonals.
The important observation\cite{Ghosh2016} is that we can write 
\begin{equation}
\label{lift}
\sum_{k=-K}^K \big( g(i+k) - g(j+k) \big)^2 = \overline{\F}(i,i) + \overline{\F}(j,j) -2 \overline{\F}(i,j).
\end{equation}
In particular, using this so-called \textit{lifting}, we can compute the patch distance using just three samples of $\overline{\F}$, one multiplication, and two additions.
The computational gain comes from the fact that the box filtering in \eqref{convn} can be computed using $O(1)$ operations with respect to $K$ using recursions\cite{Ghosh2016}.
Moreover, following the observation that not all samples of $ \overline{\F}$ are used in \eqref{NLM1D}, an efficient mechanism for computing (and storing) just the required samples was  proposed\cite{Ghosh2016}. The per-pixel complexity of computing \eqref{NLM1D} using lifting reduces to $O(S)$ from the brute-force complexity of $O(SK)$. 
Unfortunately, extending lifting to handle two-dimensional patches turns out to be difficult. Instead, we proposed to use separable filtering, where the rows (columns) are filtered using \eqref{NLM1D} followed by the columns (row). The two distinct outputs are then optimally combined to get the final image. In fact, the reason behind the averaging was to suppress artifacts in the form of stripes arising from the separable filtering. This is demonstrated with an example in Fig. \ref{example}, where we have compared  NLM, SNLM, and the proposed approach. We used bilateral filtering to remove the stripes in SNLM, at an additional cost. However, the final image still has some residual artifacts.

\section{Proposed Approach}
\label{sec:proposed}

We see less stripes in Fig. \ref{example}(d) precisely because we use a two-dimensional search. In other words, we use a cross between  classical NLM and SNLM in which we use  \eqref{proposed} for the aggregation and \eqref{weights} for the weights. The two-dimensional search results in the averaging of pixels from across rows (and columns). This does not happen in SNLM, which causes the stripes to appear in Fig. \ref{example}(c). 

\begin{figure}[!h]
\centering
 \includegraphics[width=0.75\linewidth]{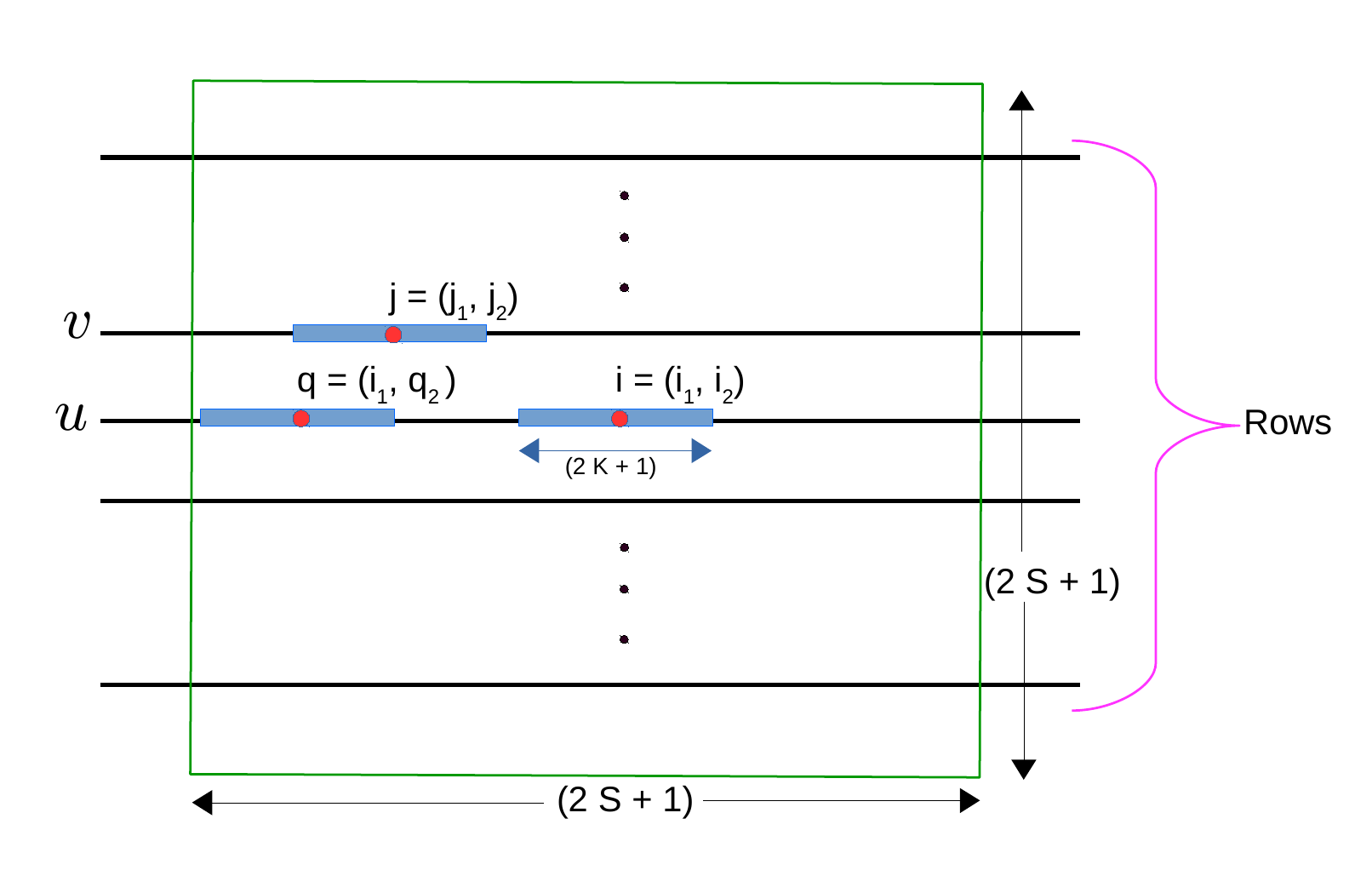}
 \caption{Illustration of the idea behind the proposed method (see text for details).}
 \label{fig:schematic}
\end{figure}

The working of our proposal is explained in Fig \ref{fig:schematic}. The pixel of interest in this case is the pixel at position $i=(i_1,i_2)$ marked with a red dot. The search window of length $2S+1$ is marked with a green bounding box. Two neighboring pixels at locations $j=(j_1,j_2)$ and $q=(i_1, q_2)$ are marked with red dots. The former pixel is on a neighboring row, while the latter is on the same row as the pixel of interest. Similar to SNLM \cite{Ghosh2016}, we can consider either horizontal or vertical patches. For our example, the patches (of length $2K+1$) are aligned with the image rows; they are marked with light blue rectangles. For our proposal, the denoising at $i=(i_1,i_2)$ is performed using the formula:
 \begin{equation}
\label{proposed}
\frac{\sum_{\ell \in S(i)} w_{i\ell} f(\ell)}{\sum_{\ell \in S(i)} w_{i\ell} },
\end{equation}
and
\begin{equation}
\label{wtsProp}
 w_{i \ell} = \exp \Big(\! - \frac{1}{\beta^2} \sum_{k=-K}^K \big( f(\ell_1,\ell_2+k) - f(i_1,i_2+k) \big)^2 \Big),
\end{equation}
where $S(i)=i+[-S,S]^2$ and $\ell=(\ell_1,\ell_2)$. To compute \eqref{proposed}, we group the neighboring patches into two categories: (i) patches with row index $i_1$, e.g., patch $q$ in Fig. \ref{fig:schematic}, and (ii) patches with a different row index, e.g., patch $j$ in the figure. 
Let $\{u(t): 1 \leq t \leq N\}$ and $\{v(t): 1 \leq t \leq N\}$ be the $i_1$-th and $j_1$-th row, where $N$ is the length of a row (see Fig \ref{fig:schematic}).  
Similar to \eqref{tensor} and \eqref{convn}, we define the $N \times N$ matrices:
\begin{equation}
\label{tensors}
\F_{uu}(p,q) = u(p) u(q), \quad \F_{vv}(p,q)=v(p)v(q), \quad \text{and} \quad \F_{uv}(p,q)=u(p)v(q),
\end{equation}
and the corresponding matrices $\overline{\F}_{uu}, \overline{\F}_{vv}$, and $\overline{\F}_{uv}$, where, for example,
\begin{equation}
\label{Fu_bar}
\overline{\F}_{uu}(p,q) =  \sum_{k=-K}^K  u(p+k) u(q+k) \qquad (1 \leq p,q \leq N).
\end{equation}
As in \eqref{lift}, the (squared) distance between patches centered at $i=(i_1,i_2)$ and $q = (i_1, q_2)$ is 
\begin{equation}
\label{liftuu}
\overline{\F}_{uu}(i_2,i_2) + \overline{\F}_{uu}(q_2,q_2) -2 \overline{\F}_{uu}(i_2, q_2).
\end{equation}
On the other hand,  the distance between patches centered at $i=(i_1,i_2)$ and $j=(j_1,j_2)$ is 
\begin{equation}
\label{liftuv}
\overline{\F}_{uu}(i_2,i_2) + \overline{\F}_{vv}(j_2,j_2) -2 \overline{\F}_{uv}(i_2,j_2).
\end{equation}
In other words, we can compute the distance between patches centered at $i$ and $q$ using $\overline{\F}_{uu}$. To compute the distance between patches centered at $i$ and $j$, we require the matrices $\overline{\F}_{uu}$, $\overline{\F}_{vv}$, and $\overline{\F}_{uv}$. Moreover, using these matrices, we can compute patch distances for different $i,j$, and $q$, provided the row index of $i$ and $q$ is $i_1$, and the row index of $j$ is $j_1$.
Thus, an efficient way of computing \eqref{proposed} is to sequentially process the rows. For each row (fixed $u$), we compute $\overline{\F}_{uu}$, $\overline{\F}_{vv}$, and $\overline{\F}_{uv}$, where $v$ corresponds to neighboring rows that are separated by at most $S$. We compute $2S+1$ matrices of the form  $\overline{\F}_{vv}$ and another $2S$ matrices of the form $\overline{\F}_{uv}$. As mentioned in Section \ref{sec:prior works}, we can compute each matrix using $O(1)$ operations with respect to $K$. Moreover, as per the sum in \eqref{NLM1D}, we only require entries within the diagonal band $\{1 \leq i,j \leq N: \  |i-j| \leq S\}$ of each matrix.
The cost of computing the banded entries is thus $O(NS)$ for each matrix. The overall cost of processing $N$ rows is $O(N^2S^2)$.  The per-pixel complexity of computing \eqref{proposed} using the proposed approach is thus $O(S^2)$. We can efficiently compute (and store) the banded entries using the method in Section 2.2 of the original paper\cite{Ghosh2016}. The main difference with SNLM is that we require a total of $4S+1$ matrices for processing each row; whereas, just one matrix is required in SNLM. As shown in Fig. \ref{example}(d), some residual noise can still be seen after the processing mentioned above. We perform a similar processing once more, except this time we use one-dimensional patches along columns. The visual quality and PSNR  of the final image (Fig. \ref{example}(e)) are comparable to NLM (Fig. \ref{example}(h)). Moreover, we see from Figs. \ref{example}(e) and \ref{example}(f) that if we first use one-dimensional patches along columns and then along rows, then the outputs are similar. 
We empirically corroborate these observations in the next section.
Therefore, we propose to first process the rows using \eqref{proposed} and then process the columns of the intermediate image using \eqref{proposed}. 
A precise description of the proposed approach for processing the (noisy) image along rows using lifting is provided in Algorithm \ref{algo1}.  We then perform column processing on the intermediate image to obtain the final output of our algorithm. That is, we simply apply Algorithm \ref{algo1} on the intermediate image, where we logically switch   the  rows and columns in the algorithm. 
Suppose $S_1$ and $S_2$ are the corresponding search windows for the row-aligned and column-aligned processing. Then we set the search parameter in Algorithm \ref{algo1} as:  $S = S_1$ for the row-aligned processing, and $S = S_2$ for the column-aligned processing.

\IncMargin{1.5mm}
\begin{algorithm}[!h]
\label{algo1}
\LinesNumbered

\KwData{Image $f$ of size $M \times N$, and parameters $K, S, \beta$.}
\KwResult{Row-processed image $\tilde{f}$ of size $M \times N$ given by \eqref{proposed}.}
\For{$i_1 = 1,\ldots, M$}{
 \% \texttt{lifting} \\
\For{$i_2 = 1, \ldots, N$}{
$u(i_2) = f(i_1, i_2)$; \\
}
Compute matrices $\F_{uu}$ and $\overline{\F}_{uu}$ using \eqref{tensors} and \eqref{Fu_bar}; \\
\For{$j_1 \in \{\ i_1 -S,\ldots, i_1 +S\} \backslash \{i_1\}$}{
\For{$j_2 = 1, \ldots, N$}{
$v(i_2) = f(j_1, j_2)$; \\
}
Compute matrices $\F_{vv}, \F_{uv}$, $\overline{\F}_{vv}$ and $\overline{\F}_{uv}$ using \eqref{tensors} and \eqref{Fu_bar}; \\
}
\% \texttt{weight computation and pixel aggregation} \\
\For{$i_2 = 1, \ldots, N$}{
Set $P = 0$ and $Q = 0$; \\
\For{$j_1 = i_1$}{
\For{$k_2 \in \{ i_2 - S,\ldots, i_2 + S\}$}{
Compute weight $w_{i_2 k_2}$ using \eqref{liftuu} and \eqref{wtsProp}; \\
$P = P + w_{i_2 k_2} f(j_1, k_2)$ ; \\
$Q = Q + w_{i_2 k_2} $; \\
}
}
\For{$j_1 \in \{\ i_1 -S,\ldots, i_1 +S\} \backslash \{i_1\}$}{
\For{$j_2 \in \{i_2 - S,\ldots, i_2 + S\}$}{
Compute weight $w_{i_2 j_2}$ using \eqref{liftuv} and \eqref{wtsProp}; \\
$P = P + w_{i_2 j_2} f(j_1, j_2)$ ; \\
$Q = Q + w_{i_2 j_2} $; \\
}
}
$\tilde{f}(i_1, i_2) = P/Q$.
}
}
\caption{Proposed processing along rows using lifting.}
\end{algorithm}
\DecMargin{1.5mm}

\begin{table}[!h]
\setlength{\tabcolsep}{2.2pt}
\setlength{\extrarowheight}{4pt}
\centering
\begin{tabular}{|c||c|c|c|c|c || c|c|c|c|c|}
\hline
$\sigma$ & 5 & 10 & 20 & 30 & 50  & 5 & 10 & 20 & 30 & 50 \\   [5pt]  \hline     \hline
Method & \multicolumn{5}{c||} {\textbf{House} ($256 \times 256$)}  &  \multicolumn{5}{c|} {\textbf{Montage} ($256 \times 256$)} \\  [5pt] \hline 
Noisy & 34.1/83  & 28.1/60   & 22.1/34    & 18.6/22   & 14.2/12  
      & 34.2/83  & 28.1/61   & 22.1/36    & 18.6/25   & 14.1/15  \\ [5pt]    \hline 
NLM \cite{Buades2005}   & 36.9/90  & 34.1/87   & 29.7/82    & 26.8/77   & 24.0/69 
      & 39.1/97  & 34.3/94   & 29.6/89    & 26.5/85   & 22.2/76  \\  [5pt]   \hline
Darbon et al. \cite{Darbon2008}   & 36.1/90  & 31.4/75   & 26.1/51    & 22.8/36   & 18.6/21 
                                  & 38.4/89  & 30.9/76   & 25.8/52    & 22.6/38   & 18.6/24  \\  [5pt]   \hline
SNLM \cite{Ghosh2016} & 36.6/89  & 33.6/86   & 29.4/81    & 26.5/76   & 23.7/69
      & 39.3/97  & 34.6/94   & 29.7/89    & 26.7/84   & 22.5/77   \\  [5pt]   \hline
Proposed & 36.6/89  & 34.1/86  & 30.4/82  & 27.3/77   & 24.1/70 
	 & 39.4/97  & 34.8/94  & 30.2/90  & 27.3/86   & 23.4/79  \\     [5pt]     \hline
BM3D \cite{BM3D}   & 38.6/95  & 34.7/93  & 31.3/88  & 29.3/85  & 26.6/78 
	 & 41.1/98  & 37.3/96  & 33.5/94  & 31.2/91   & 27.4/85  \\     [5pt]     \hline
	 \hline
\textit{Method} & \multicolumn{5}{c||} {\textbf{Boat} ($512 \times 512$)} & \multicolumn{5}{c|} {\textbf{Man} ($1024 \times 1024$)} \\ \hline
Noisy & 34.1/97   & 28.1/90   & 22.1/73   & 18.6/59   & 14.2/41 
      & 34.1/99   & 28.1/97   & 22.1/91   & 18.6/85   & 14.1/70 \\   [5pt]  \hline 
NLM \cite{Buades2005}  & 35.1/96   & 30.8/89   & 26.7/78   & 24.7/70   & 23.0/62 
      & 35.3/98   & 31.1/95   & 27.5/88   & 25.8/83   & 24.2/76 \\   [5pt]  \hline
Darbon et al. \cite{Darbon2008}   & 34.4/97  & 30.3/94   & 25.4/82    & 22.4/71   & 18.4/53 
				  & 35.1/97  & 30.4/98   & 25.6/95    & 22.5/90   & 18.5/80  \\  [5pt]   \hline
SNLM \cite{Ghosh2016} & 35.0/96   & 30.7/89   & 26.6/77   & 24.5/69   & 22.7/61
      & 35.3/98   & 31.0/95   & 27.2/87   & 25.4/83  & 23.8/75 \\  [5pt]   \hline
Proposed  & 34.9/96  & 30.7/89  & 26.8/77   &  24.7/70  & 22.9/62 
          & 35.1/98  & 31.1/95  & 27.5/88   &  25.8/ 84 & 24.2/78 \\  [5pt]   \hline
BM3D \cite{BM3D} & 37.3/98  & 33.9/96  & 30.8/92  & 29.0/88   & 26.7/81 
	 & 37.3/99  & 34.1/98  & 31.2/96  & 29.5/94   & 27.4/89  \\     [5pt]     \hline
	 \hline 
\end{tabular}
 \\~\\
\caption{Comparison of the denoising performances on various images \cite{ImgDatabase1} in terms of PSNR/SSIM at various noise standard deviations $\sigma$. The PSNRs are rounded to one decimal place, while the SSIMs (in $\%$) are rounded to  integer. }
\label{tab:psnr}
\end{table}
\begin{table}[!htp]
\setlength{\tabcolsep}{4.0pt}
\setlength{\extrarowheight}{3pt}
\centering
\begin{tabular}{|c||c|c|c || c|c|c|}
\hline
$S$ & 7 & 10 & 12   & 7 & 10 & 12  \\   [5pt]  \hline     \hline
Method & \multicolumn{3}{c||} {$K = 2$}  &  \multicolumn{3}{c|} {$K = 3$}
\\  [5pt] \hline
NLM  \cite{Buades2005}    	 & 44  & 87   & 124
				 & 45  & 88   & 126     \\ [5pt]    \hline
Darbon et al. \cite{Darbon2008}   & 0.33  & 0.60   & 0.84      
				 & 0.33  &  0.62 & 0.85  \\ [5pt]    \hline
SNLM \cite{Ghosh2016}     & 0.31  & 0.39   & 0.45      
				 & 0.32  &  0.40 & 0.46  \\ [5pt]   \hline
Proposed & 1.20  & 2.30   & 3.20
         & 1.30  & 2.40   & 3.30     \\ [5pt]    \hline
\end{tabular}
 \\~\\
\caption{Comparison of the run-time (in seconds) of the proposed approach with classical NLM  for a $256 \times 256$ image. 
The computations were performed using Matlab on a 3.40 GHz Intel quad-core machine with 32 GB memory.}
\label{tab:timing}
\end{table}
\begin{figure}[!h]
\centering
\includegraphics[width=0.97\linewidth]{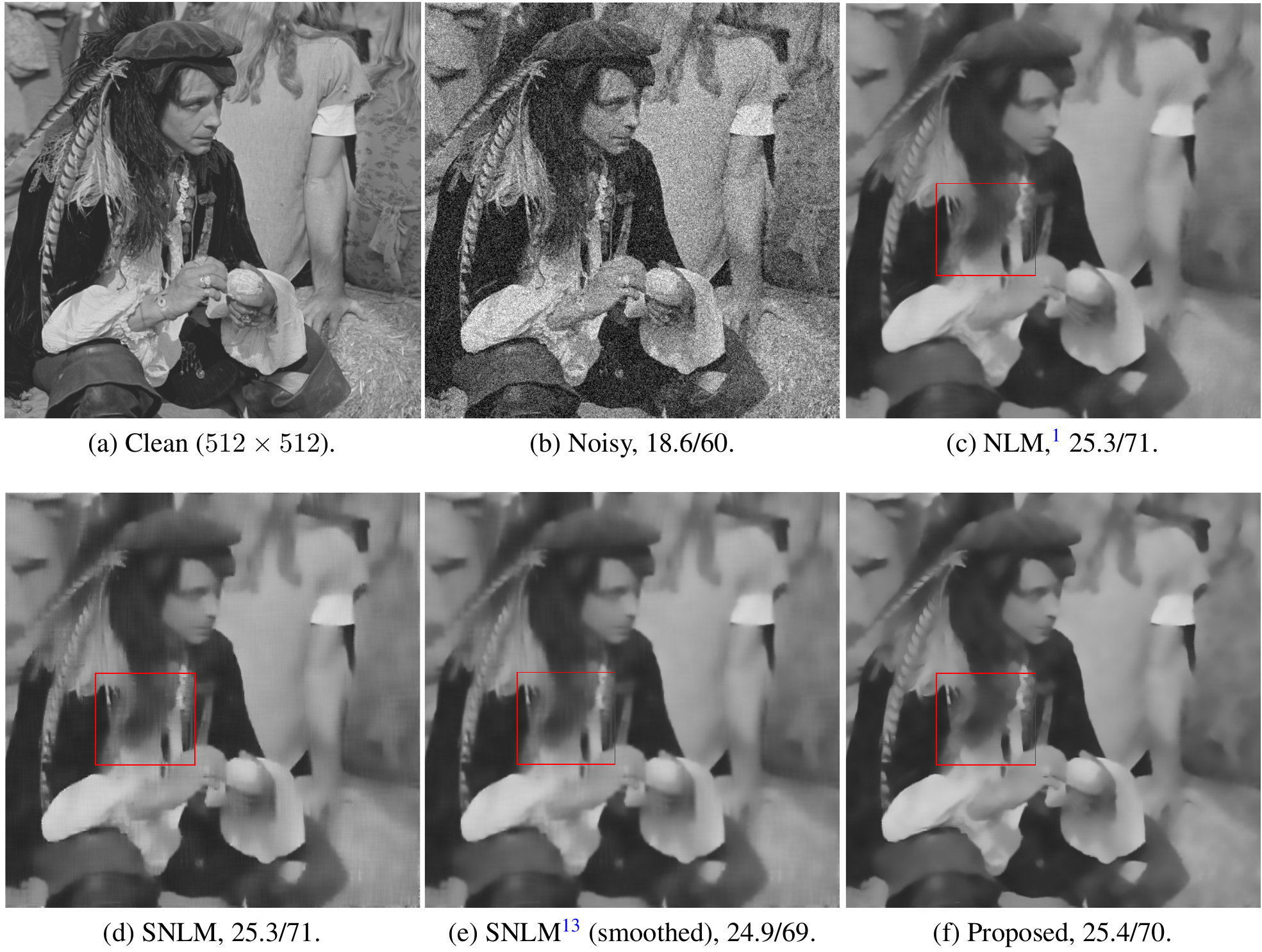}
\caption{Denoising of \emph{Man} \cite{ImgDatabase1} at $\sigma = 30$. Notice that stripes can be seen in (e) after smoothing (d) using a bilateral-filter. The PSNR/SSIM values with reference to the clean image are also provided. The result from our proposal (f) is visually similar to classical NLM (c). The runtime for NLM, SNLM, and the proposed method are 335, 1.7, and 9.8 seconds. We used the parameter settings mentioned in the main text. The PSNR/SSIM between the proposed approximation (f) and the classical NLM (c) are $40.58 / 70.22$, whereas the values are $37.96 / 69.90$ for SNLM (d).} 
\label{fig3}
\end{figure}
\begin{figure*}[!h]
\centering
\includegraphics[width=0.97\linewidth]{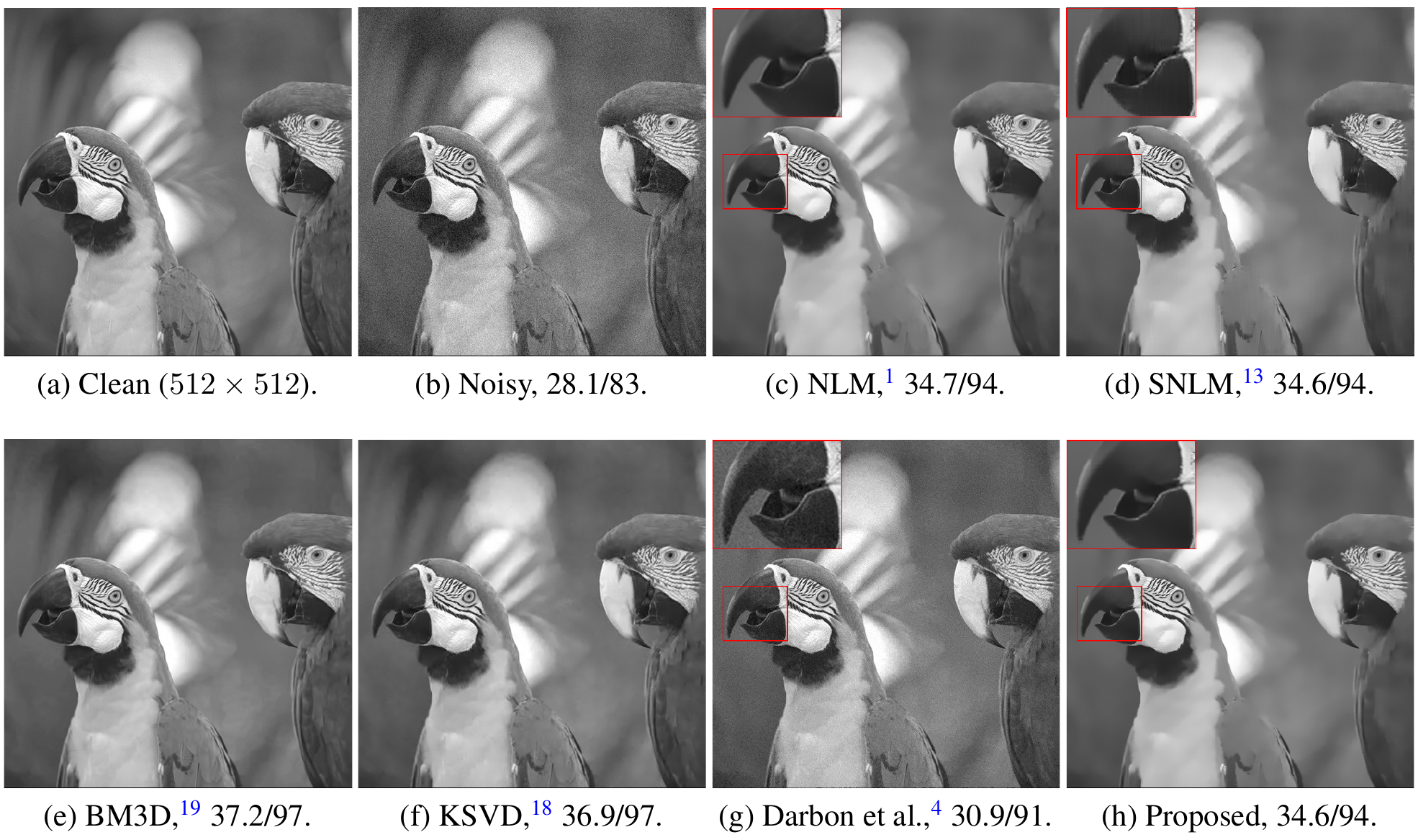}
\caption{Denoising of \emph{kodim23} \cite{ImgDatabase2} at $\sigma = 10$. 
The result obtained through our proposal (h) is visually similar to classical NLM (c). The runtime for NLM, SNLM, and the proposed method are 335, 1.7, and 9.8 seconds. 
The PSNR/SSIM between the proposed approximation (h) and the classical NLM (c) are $43.33 / 99.7$, whereas these values are $42.4/99.4$ for (d) and $30.85/89.3$ for (g).
We have zoomed the region around the beak in (c), (d), (g), and (h). We can see some artifacts in (d) and residual noise in (g); the zooms in (c) and (h) are visually indistinguishable.
} 
\label{fig4}
\end{figure*}

\section{Experiments}
\label{sec:results}

The denoising performance of the proposed method is compared with NLM and SNLM in Table \ref{tab:psnr}. 
We have used standard grayscale images from \cite{ImgDatabase1, ImgDatabase2} for our experiments.
The Matlab implementation used to generate the results in this section is publicly available\footnote{\url{http://in.mathworks.com/matlabcentral/fileexchange/64856}}.
The search windows for the three methods were set as follows. Suppose $S$ be the search window for NLM (which we take as reference). Following the original proposal\cite{Ghosh2016}, the window for SNLM is also set as $S$.
For a fair comparison with NLM, we ensure that equal number of pixel are averaged in both methods. This is achieved if $(2 S_1 +1)^2 + (2 S_2 + 1)^2 = (2 S + 1)^2$. Moreover, following\cite{Gastal2011}, we set  $S_2 = S_1/2$. These equations uniquely determine $S_1$ and $S_2$ (up to an integer rounding).
Moreover, we normalize the smoothing parameters in \eqref{wtsNLM} and \eqref{wtsProp} using the relation $\beta^2 =\alpha^2/(2K+1)$. For the results in   Table \ref{tab:psnr}, we set $K = 3$, $S = 10$, $S_1 = 9$, $S_2 = 4$, and $\alpha = 10 \sigma$. We notice from Table \ref{tab:psnr} that the proposed approach gives comparable results in terms of PSNR and SSIM \cite{SSIM2004}. A visual comparison of the denoising results is provided in Fig. \ref{fig3} and \ref{fig4}. We can clearly see some stripes in the images obtained using SNLM, both with and without post-processing (see the boxed areas). In contrast, there is hardly any artifacts present in the denoised image obtained using our method. A timing comparison is provided in Table \ref{tab:timing}. While the proposed method is slower than SNLM (this is the price we pay for removing the stripes), it is nevertheless significantly faster than NLM. 

We note that though Darbon et al. \cite{Darbon2008} is generally faster than our current proposal, its denoising performance starts deteriorating with the increase in noise variance. This is evident from Table \ref{tab:psnr} and Fig. \ref{fig4}. We also note that NLM and SNLM fall short of KSVD \cite{KSVD} and BM3D \cite{BM3D} in terms of denoising performance. Nevertheless, NLM continues to be of interest  due to its decent denoising capability\cite{Batikian2014, Chan2014, Chen2015, Zeng2016}, and importantly, the availability of  fast approximations. As reported by other authors\cite{Treece2016}, NLM is quite effective in preserving fine details, while successfully removing noise.

\section{Conclusion}
\label{sec:conclusion}

We proposed a method that uses the idea of lifting from previous work\cite{Ghosh2016} to perform fast non-local means denoising of images.
The proposed method does not give rise to undesirable artifacts (as was the case with the original proposal), and produces images 
whose denoising quality and PSNR/SSIM are comparable to non-local means. 
While this comes at the expense of added computation, the proposed method nevertheless is much faster than non-local means. 
In fact, the speedup is about $40$x for practical parameter settings. 

\section{Acknowledgements}

The last author was supported by a Startup Grant from IISc and EMR Grant SB/S3/EECE/281/2016 from DST, Government of India. 

\bibliographystyle{IEEEbib}

\begin{thebibliography}{9}

\bibitem{Buades2005} A. Buades, B. Coll, and J.-M. Morel, ``A non-local algorithm for image denoising," \textit{Proc. IEEE Conference on Computer Vision and Pattern Recognition}, {\bf2}, pp. 60-65 (2005).

\bibitem{Sapiro2005} M. Mahmoudi and G. Sapiro, ``Fast image and video denoising via nonlocal means of similar neighborhoods,'' \textit{IEEE Signal Processing Letters}, {\bf12}(12), pp. 839-842 (2005).

\bibitem{Wang2006} J. Wang, Y. Guo, Y. Ying, Y. Liu, and Q. Peng, ``Fast non-local algorithm for image denoising,'' \textit{Proc. IEEE International Conference on Image Processing}, pp. 1429-1432 (2006).

\bibitem{Darbon2008} J. Darbon,  A. Cunha,  T. F. Chan, S. Osher,  and G. J. Jensen, ``Fast nonlocal filtering applied to electron cryomicroscopy,''  \textit{Proc. IEEE International Symposium on Biomedical Imaging}, pp. 1331-1334 (2008).

\bibitem{Dauwe2008} A. Dauwe, B. Goossens, H. Luong, and W. Philips, ``A fast non-local image denoising algorithm,'' \textit{Proc. SPIE Electronic Imaging}, {\bf 68}(12), pp. 1331-1334 (2008).

\bibitem{Orchard2008} J. Orchard, M.  Ebrahimi, and A. Wong, ``Efficient nonlocal-means denoising using the SVD,'' \textit{Proc. IEEE International Conference on Image Processing}, pp. 1732-1735 (2008).

\bibitem{KUD2009} V. Karnati, M. Uliyar, and S. Dey, ``Fast non-local algorithm for image denoising,'' \textit{Proc. IEEE International Conference on Image Processing}, pp. 3873-3876 (2009).

\bibitem{Condat2010} L. Condat, ``A simple trick to speed up and improve the non-local means,'' \textit{Research Report}, HAL-00512801, (2010).

\bibitem{N1981} P. M. Narendra, ``A separable median filter for image noise smoothing,'' \textit{IEEE Transactions on Pattern Analysis and Machine Intelligence}, {\bf3}, pp. 20-29 (1981).

\bibitem{Fukushima2015} N. Fukushima, S. Fujita, and Y. Ishibashi, ``Switching dual kernels for separable edge-preserving filtering,'' \textit{IEEE International Conference on Acoustics, Speech and Signal Processing}, (2015).

\bibitem{Pham2005} T. Q. Pham and L. J. Van Vliet, ``Separable bilateral filtering for fast video preprocessing,'' \textit{Proc. IEEE International Conference on Multimedia and Expo}, (2005).

\bibitem{KLCLKK2011} Y. S. Kim, H. Lim, O. Choi, K. Lee, J. D. K. Kim, and C. Kim, ``Separable bilateral non-local means,'' \textit{Proc. IEEE International Conference on Image Processing}, pp. 1513-1516 (2011).

\bibitem{Ghosh2016} S. Ghosh and K. N. Chaudhury, ``Fast separable nonlocal means,'' \textit{SPIE Journal of Electronic Imaging}, {\bf25}(2),  023026 (2016).

\bibitem{Gastal2011} E. S. Gastal and M. M. Oliveira. ``Domain transform for edge-aware image and video processing,'' \textit{ACM Transactions on Graphics} (ToG), {\bf30}(4), 69 (2011).

\bibitem{ImgDatabase1} BM3D Image Database,  \url{http://www.cs.tut.fi/~foi/GCF-BM3D}.

\bibitem{ImgDatabase2} KODAK Image Database, \url{http://r0k.us/graphics/kodak/}.

\bibitem{SSIM2004} Z. Wang, A. C. Bovik, H. R. Sheikh, and E. P. Simoncelli, ``Image quality assessment: From error visibility to structural similarity,'' \textit{IEEE Transactions on Image Processing}, {\bf 13}(4), pp. 600-612 (2004).

\bibitem{KSVD} M. Elad and M. Aharon, ``Image denoising via sparse and redundant representations over learned dictionaries,'' \textit{IEEE Transactions on Image Processing}, \textbf{15}(12), pp. 3736-3745 (2006).

\bibitem{BM3D} K. Dabov, A. Foi, V. Katkovnik, and K. Egiazarian, ``Image denoising by sparse 3-D transform-domain collaborative filtering,'' \textit{IEEE Transactions on Image Processing}, \textbf{16}(8), pp. 2080-2095  (2007).

\bibitem{Batikian2014} J. M. Batikian and M. Liebling, ``Multicycle non-local means denoising of cardiac image sequences,''  \textit{IEEE International Symposium on Biomedical Imaging}, pp. 1071-1074 (2014).

\bibitem{Chan2014} C. Chan, R. Fulton, R. Barnett, D.D. Feng, and S. Meikle, ``Post-reconstruction nonlocal means filtering of whole-body PET with an anatomical prior,'' \textit{IEEE Transactions on Medical Imaging}, \textbf{33}(3), pp. 636-650 (2014).

\bibitem{Chen2015} G. Chen, P. Zhang, Y. Wu, D. Shen, and P.T. Yap, ``Collaborative non-local means denoising of magnetic resonance images,'' \textit{IEEE International Symposium on Biomedical Imaging}, pp. 564-567 (2015).

\bibitem{Zeng2016} D. Zeng, J. Huang, H. Zhang, Z. Bian, S. Niu, Z. Zhang, Q. Feng, W. Chen, and J. Ma, ``Spectral CT image restoration via an average image-induced nonlocal means filter,'' \textit{IEEE Transactions on Biomedical Engineering}, \textbf{63}(5), pp. 1044-1057 (2016).

\bibitem{Treece2016} G. Treece, ``The bitonic filter: linear filtering in an edge-preserving morphological framework,'' \textit{IEEE Transactions on Image Processing}, \textbf{25}(11), pp. 5199-5211 (2016).

\end{thebibliography}

\end{document}